\documentclass[10pt,twocolumn,letterpaper]{article}

\usepackage[pagenumbers]{cvpr} 

\usepackage{currfile} %
\usepackage{algorithm}
\usepackage{algpseudocode} %

\usepackage{xcolor}
\usepackage{colortbl}
\usepackage{tabularx}

\newcommand{\qaskdelta}[2]{%
  \mbox{#1{\fontsize{6}{7}\selectfont\raisebox{-0.35ex}{\,#2}}}%
}

\usepackage{microtype}

\renewcommand{\paragraph}[1]{\vspace{.5em}\noindent\textbf{#1.}}

\definecolor{cvprblue}{rgb}{0.21,0.49,0.74}
\usepackage[pagebackref,breaklinks,colorlinks,allcolors=cvprblue]{hyperref}

\def\confName{CVPR}
\def\confYear{2027}

\usepackage{booktabs}
\usepackage{array}
\usepackage{multirow}
\usepackage{graphicx}
\usepackage{xcolor}
\usepackage{colortbl}

\title{QASK-Edit: Adaptive Sparse Attention Across Queries and Key Sources for Efficient Video Editing}

\author{
Hengrui Liu \quad Wenbo Xu \quad Ziqi Sheng \quad Wei Lu \\
School of Computer Science and Engineering, Sun Yat-sen University, Guangzhou, China \\
}

\ifdefined\pdfsuppressptexinfo
\fi

\ifdefined\pdfinfoomitdate
\fi

\begin{document}
\maketitle
\begin{abstract}

Diffusion Transformers (DiTs) have achieved strong performance in generation based video editing, but suffer from high inference latency due to the long token sequences formed by conditioning inputs and generated videos. Although training-free sparse attention reduces this cost, existing methods typically apply a fixed retention ratio or a single threshold shared across all tokens. In this work, we view video editing as conditional generation and identify a key insight: attention demand varies significantly both across different key sources (e.g., source videos, reference images, audio, text, and generated videos) and across queries within each source. Motivated by this, we propose \textbf{QASK-Edit} (\textbf{Q}uery-\textbf{A}daptive \textbf{S}parsity over \textbf{K}ey Sources for Video Editing), a training-free sparse attention framework that allocates computation independently for each query and key source. Specifically, for each query, QASK-Edit considers both the attention mass and approximation risk of each block. It then evaluates candidate routing thresholds against each key source's risk budget, choosing exact computation or approximations to dynamically adjust sparsity for each query. Extensive experiments on three video editing models demonstrate that QASK-Edit achieves up to \(1.46\times\) end-to-end speedup over dense attention pipeline while preserving both editing quality and unedited region fidelity. Furthermore, QASK-Edit generalizes directly to Reference-to-Video generation, delivering up to \(1.36\times\) speedup while maintaining generation quality.

\end{abstract}    
\section{Introduction}
\label{sec:intro}

Combining diffusion models~\cite{ho2020denoising} with attention mechanisms, diffusion transformers (DiTs)~\cite{peebles2023scalable} have enabled powerful video foundation models~\cite{wan2025wan} and opened new possibilities for video editing. As video DiTs continue to advance, editing methods increasingly reuse pretrained video generators or adapt video foundation models to perform editing directly~\cite{ku2024anyv2v,shao2026liveditor}. This generation-based formulation supports diverse editing operations, but also inherits the high attention cost of the underlying models. In particular, jointly attending to input video context and the video being edited further increases sequence length, amplifying the quadratic cost of attention~\cite{shao2026liveditor}. Efficient attention that preserves both the intended edits and unaffected content is therefore essential for practical video editing.

\begin{figure*}[t]
\centering
\includegraphics[width=\linewidth]{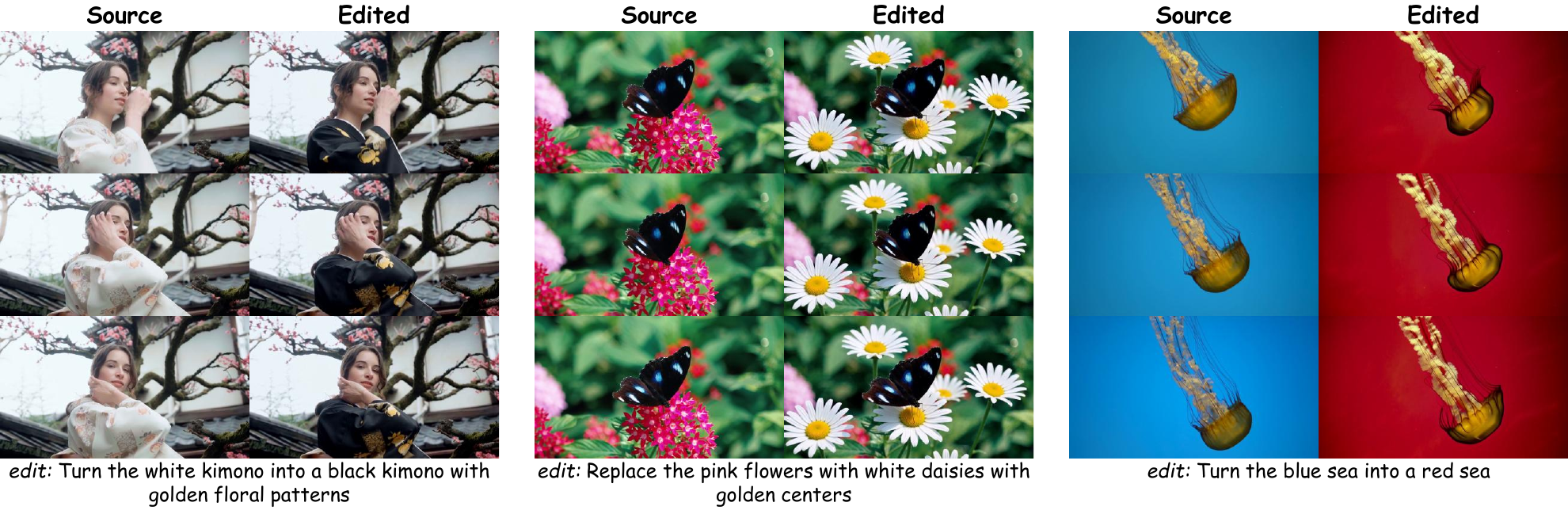}
\vspace{-1.3em}
\caption{Qualitative video editing results of QASK-Edit on MiniMax-H3. Three sampled frames are shown for each source–edited video pair, covering diverse editing tasks such as clothing modification, object replacement, and scene recoloring. The results demonstrate high-quality edits with preserved temporal consistency.}
\label{fig:visual_glass}
\vspace{-1em}
\end{figure*}

A large body of work~\cite{zhang2025sta,xu2025xattention,yang2025svg2,liu2026spade,li2026solattn,zhou2026svgear,shao2026liveditor} reduces this cost through sparse attention, with training-free methods requiring no updates to model weights. Pattern-based approaches exploit localized spatiotemporal structure to restrict attention computation~\cite{zhang2025sta}. Content-adaptive approaches use lightweight importance scores to select blocks~\cite{xu2025xattention}. For block formation, SVG2~\cite{yang2025svg2} clusters and reorders semantically similar tokens, while SPADE~\cite{liu2026spade} selects spatial, temporal, or mixed partitions according to intra-block feature similarity. Hybrid approaches combine sparse exact computation with low-cost approximations: Sol-Attn~\cite{li2026solattn} uses query-dependent thresholds calibrated by a shared standardized cutoff \(\beta\), rather than fixed block counts, and reuses token-to-block scores for routing and pooled approximation under a shared softmax. SVG-EAR~\cite{zhou2026svgear} couples parameter-free centroid compensation with error-aware routing based on the estimated error-to-cost ratio. In video editing, LIVEditor~\cite{shao2026liveditor} introduces In-Context Sparse Attention (ISA), combining context pre-selection with query grouping based on attention sharpness. Within the selected context, sharp queries use full attention, while flatter queries combine exact computation with pooled key/value approximations. While these methods reduce attention cost, preserving editing quality also depends on how well their sparsity decisions match the conditioning requirements of individual queries.

We revisit generation-based video editing from the perspective of conditional video generation. For each query, keys from the input video, editing instructions, audio, and the video being generated serve as distinct conditioning sources. Existing sparse attention methods, however, usually control the extent of exact computation either with a prescribed block count or retention ratio, or with a selection criterion shared across all sources. For example, ISA~\cite{shao2026liveditor} pre-selects a fixed fraction of context blocks before query-adaptive computation, whereas Sol-Attn~\cite{li2026solattn} allows query-dependent block counts but calibrates them with a globally shared threshold. Such designs implicitly assume that a common sparsity rule can serve different conditioning sources and queries. Our preliminary analysis in Sec.~\ref{sec:qask_motivation} challenges this assumption: attention requirements differ substantially across key sources and also vary considerably within each source over query tiles, heads, layers, and denoising steps. These observations motivate \emph{query-adaptive sparsity over key sources}, which independently adapts the sparsity of each source to the current query, preserving necessary conditioning information while avoiding unnecessary exact computation.

Motivated by these observations, we propose \textbf{QASK-Edit} (Query-Adaptive Sparsity over Key Sources for Video Editing), a training-free sparse attention framework. Following SPADE's similarity-guided partitioning~\cite{liu2026spade}, we group similar tokens into blocks and construct compact, representative summaries. For each query tiles, QASK-Edit adapts sparsity independently for each key source according to approximation risk. Inspired by former work, we connect attention sharpness with approximation error~\cite{zhou2026svgear, shao2026liveditor}, thus our risk proxy combines each block's estimated share of attention mass with the dispersion of its sub-block scores. Each source adopts the most aggressive sparsity level that satisfies its risk budget. Considering intra-block score dispersion allows us to distinguish coherent, broadly relevant regions from heterogeneous blocks containing a few salient interactions, guiding the allocation of exact and approximate computation. The resulting attention outputs feed the pretrained editor's original denoising pipeline, whose final video latents are decoded into the edited video.

Our experiments demonstrate the efficiency and transferability of {QASK-Edit}. \textbf{First}, compared with sparse attention methods used for video generation and ISA~\cite{shao2026liveditor} designed for generation-based video editing, QASK-Edit achieves the highest end-to-end speedups among the evaluated baselines across three editing models. QASK-Edit delivers up to \(1.46\times\) end-to-end speedup over the released dense-attention pipelines while retaining competitive editing quality on FiVE-Bench~\cite{li2025five} and IVE-Bench~\cite{chen2026ivebench}. \textbf{Second}, ablation studies validate the benefits of distinguishing key sources and adapting sparsity to individual queries. \textbf{Finally}, we apply the same framework to image-conditioned video generation with VACE~\cite{jiang2025vace} and Phantom~\cite{liu2025phantom}, achieving up to \(1.36\times\) end-to-end speedup with quality comparable to dense attention. These results demonstrate its potential for broader reference-conditioned generation tasks.

Our main contributions are summarized as follows:
\begin{itemize}
    \item We revisit video editing as reference-conditioned generation and identify heterogeneous attention requirements across key sources and within each source, revealing the limitations of uniform fixed-count or fixed-ratio block selection.

    \item We propose {QASK-Edit}, a training-free sparse attention framework that adapts computation to each query and key source. Guided by approximation risk estimated from attention mass and sub-block score dispersion, it selectively applies pooled approximation, exact computation, or finer-grained routing within a block hierarchy.

    \item We demonstrate effective acceleration with competitive editing quality across three video editing models and validate transferability to two image-conditioned video generation models, supporting the applicability of QASK-Edit to broader reference-conditioned generation tasks.
\end{itemize}
\section{Related Work}
\label{sec:related_work}

\subsection{Video Editing}

Video editing methods extend image diffusion priors through per-video adaptation~\cite{wu2023tuneavideo}, attention and feature control~\cite{qi2023fatezero,liu2024videop2p,geyer2024tokenflow,cong2024flatten}, or temporally coordinated sampling and propagation~\cite{khachatryan2023text2videozero,yang2023rerender,kara2024rave}. More recent approaches reuse pretrained video generators~\cite{ku2024anyv2v,jiao2026unieditflow,jiao2026streamedit}, train instruction-guided editors on paired data~\cite{bai2025ditto,mai2026easyv2v}, and support unified generation and editing through contextual conditioning~\cite{jiang2025vace,shao2026liveditor,minimax2026h3}. Source videos and additional conditions guide the generated content, but their attention interactions also increase computational cost. Our work focuses on accelerating these interactions within pretrained editors while preserving editing quality.

\subsection{Sparse and Approximate Video Attention}

FlashAttention~\cite{dao2022flashattention} improves exact attention through IO-aware execution. Sparse methods further reduce evaluated interactions using spatiotemporal patterns~\cite{zhang2025sta,li2025radial}, content-dependent block selection~\cite{xu2025xattention,zhang2025spargeattention,zhang2025vsa}, or editing-specific motion priors~\cite{mahmud2025adave}. Semantic clustering and adaptive partitioning improve block representations for routing~\cite{yang2025svg2,liu2026spade,luo2026svoo}. Hybrid approaches retain approximate contributions alongside sparse exact computation, including trainable sparse--linear attention~\cite{zhang2025sla,zhang2026sla2} and parameter-free pooled compensation~\cite{li2026solattn,zhou2026svgear}. These approaches motivate considering both block relevance and approximation reliability when allocating computation.

Sol-Attn~\cite{li2026solattn} permits query-dependent exact-computation budgets through score thresholds calibrated by a shared standardized cutoff. LIVEditor's ISA~\cite{shao2026liveditor} distinguishes context and source tokens, combining fixed-ratio context pre-selection with sharpness-based query grouping and pooled approximation. SVG-EAR~\cite{zhou2026svgear} estimates centroid-compensation error and selects blocks by their error-to-cost ratio under a prescribed density budget, explicitly separating attention mass from approximation difficulty. In {QASK-Edit}, source-specific risk budgets guide independent threshold calibration for each query tile and key source, allowing their exact-computation budgets to vary with approximation risk. We estimate this risk using attention-mass-weighted sub-block score dispersion. Hierarchical routing further permits selective sub-block refinement, retaining coarse approximations where reliable while resolving local interactions hidden by block pooling.

\section{Method}
\label{sec:method}

We introduce \textbf{QASK-Edit}, a training-free attention framework for generation-based video editing that allocates computation separately for each query tile and key source, following our main insight. For the current query tile, QASK-Edit selects a sparsity threshold for each key source, allowing computation allocation to vary both across sources and with query demand within each source. We use hierarchical block summaries to estimate approximation risk and select these thresholds under source-specific risk budgets. Queries within a tile share a routing pattern, while exact and approximate contributions are aggregated separately for each original query under a shared softmax.

\begin{figure*}[!t]
\centering
\includegraphics[width=\linewidth]{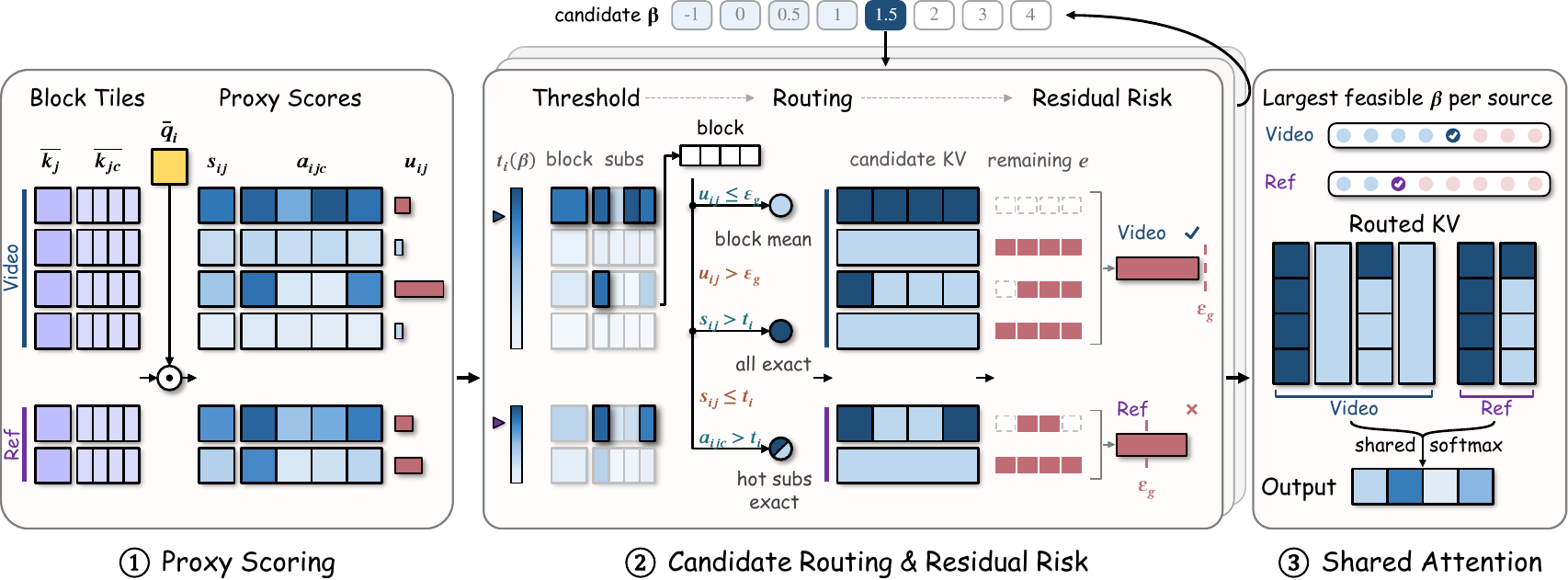}
\vspace{-1.3em}
\caption{
\textbf{Overview of QASK-Edit.} The process begins with the generation of summaries for blocks, sub-blocks, and second-level sub-blocks. For each query tile, we independently select a routing threshold for each key source within the risk budget, based on the residual risk introduced by approximate calculations. This routing threshold then determines whether to perform exact or approximate calculations. Finally, we aggregate the contributions of all key sources into a single softmax and output the final result.}
\label{fig:qask_overview}
\vspace{-0.9em}
\end{figure*}

\subsection{Motivation}
\label{sec:qask_motivation}

We revisit video editing from the perspective of conditional video generation. Whether or not inversion is required, video editing can be viewed as generating a new video from a set of contextual inputs. These include a representation of the source video, such as its latent representation or an interpolation into the noise space. The context also includes the video being generated and, when available, editing instructions, audio, and other inputs. Attention to the video being generated may be viewed as self-attention in a broad sense. We focus primarily on methods that retain the source video as a reference during generation. Each attention query therefore attends to keys from different sources. We hypothesize that the proportion of the most relevant attention interactions differs across key sources. The model's focus on these sources may also change over the course of generation. For example, it may focus more on the source video at the beginning of editing and more on the new video during later refinement.

To examine whether a common sparsity rule suits generation-based video editing, we analyze the attention requirements of individual key sources in StreamEdit~\cite{jiao2026streamedit} and MiniMax-H3~\cite{minimax2026h3}. For each combination of query tile, head, layer, and denoising step, we rank blocks within each source by their total dense attention weight. We then record the minimum block count and percentage needed to cover \(90\%\) of that source's attention mass. We report the mean requirements and the average within-task standard deviation. Each standard deviation is computed across heads, layers, and denoising steps, with both the query tile and key source fixed.

\begin{figure}[t]
\centering
\includegraphics[width=0.95\linewidth]{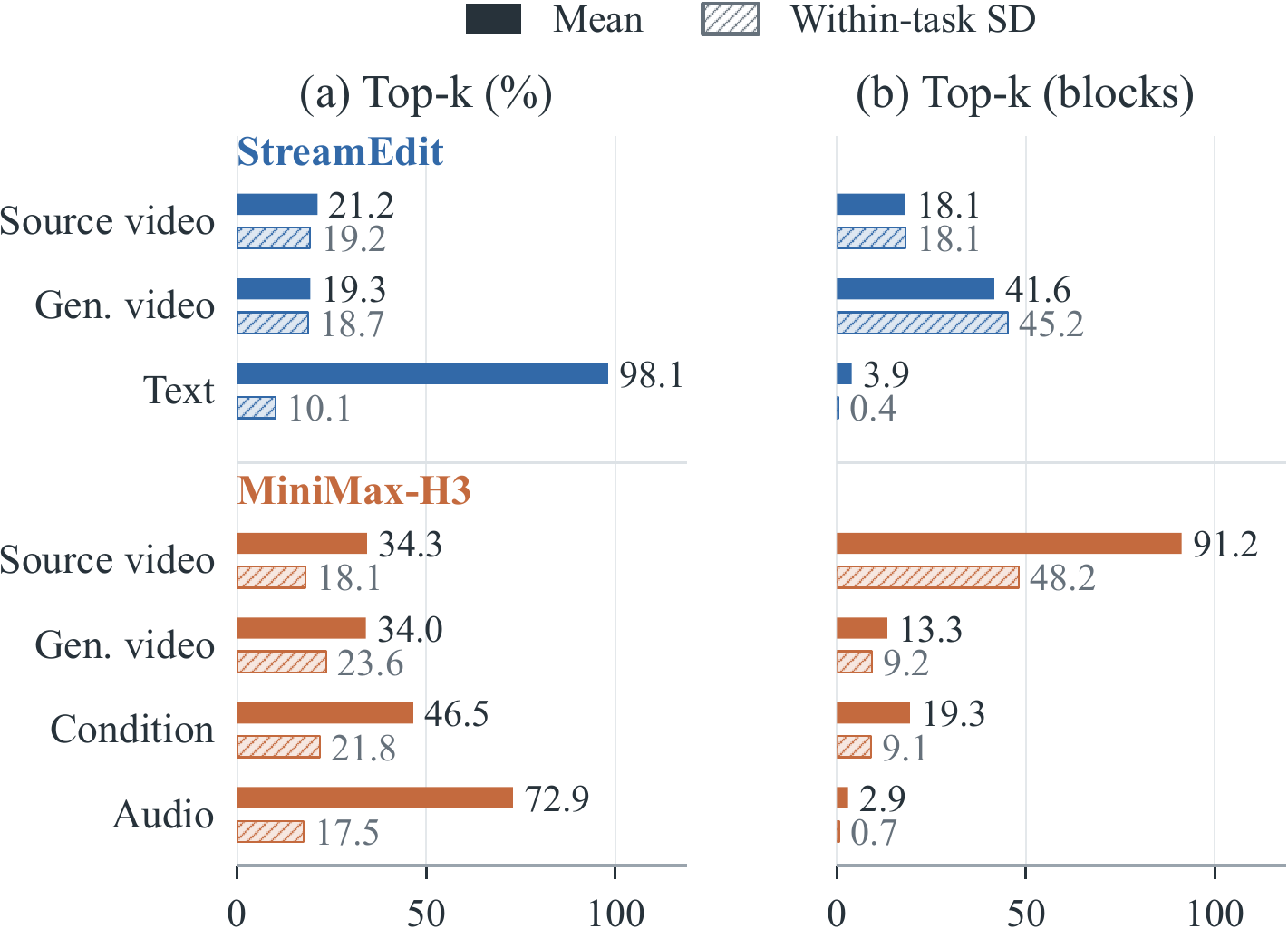}
\caption{\textbf{Attention requirements vary across key sources and within each source.} For each source, we report the mean minimum block count and percentage required to cover \(90\%\) of its dense attention mass, along with their average within-task standard deviations.}
\label{fig:key_source_sparsity}
\vspace{-1em}
\end{figure}

Figure~\ref{fig:key_source_sparsity} confirms our hypothesis. First, the mean requirements differ across key sources in both absolute block count and relative percentage. Second, for a fixed query tile and key source, the within-task standard deviation is often comparable to the corresponding mean. These observations motivate source-specific sparsity decisions that adapt to the current query tile. To choose sparsity levels under these changing attention requirements, we introduce an approximation-risk proxy in Sec.~\ref{sec:qask_routing}. This proxy helps determine where each type of computation is needed. We further evaluate the benefits of query-adaptive allocation over source-fixed thresholds in Sec.~\ref{sec:ablation_study}.

\subsection{Representative Blocks}
\label{sec:qask_blocks}

This section defines the quantities used throughout the method. We describe the algorithm within a single attention operation. Let \(i\) index a query tile and \(j\) a KV block. Each block contains four disjoint sub-blocks indexed by \(c\). We use \(f\) to index the second-level sub-blocks introduced below. At each level, indices are unique within the current attention operation. Each sub-block has a unique parent, so parent indices are omitted. We use \(g(j)\) to denote the source of block \(j\).

Following SPADE~\cite{liu2026spade}, we use the Sum of Intra-block Cosine Similarities (SICS) to choose block layouts independently for queries and each video key source. The resulting blocks can be spatial, temporal, or spatiotemporal. This choice favors internally coherent blocks and makes their summaries more representative. Values follow the same partition as their corresponding keys. Other modalities use partitions suited to their own structure. We compute mean query, key, and value representatives for each block, and mean key and value representatives for sub-blocks formed along the selected layout. The block and sub-block relevance scores are
\begin{equation}
s_{ij}=\frac{\bar q_i^{\top}\bar k_j}{\sqrt d},
\qquad
a_{ic}=\frac{\bar q_i^{\top}\bar k_c}{\sqrt d},
\label{eq:qask_scores}
\end{equation}
where \(d\) is the head dimension. The block score summarizes overall relevance. The sub-block scores reveal local responses that the block mean may hide. Let \(n_j\) and \(n_c\) denote the valid token counts of block \(j\) and sub-block \(c\), respectively. The valid sub-block set \(\mathcal C_j\) contains the sub-blocks of \(j\) with \(n_c>0\). For \(c\in\mathcal C_j\), we define the sub-block token fraction as \(w_c=n_c/n_j\).

For risk estimation, we further divide each sub-block into two disjoint second-level sub-blocks. Their mean keys and valid token counts are denoted by \(\bar k_f\) and \(n_f\), respectively. Let \(\mathcal F_c\) contain the second-level sub-blocks of \(c\) with \(n_f>0\), so that \(\sum_{f\in\mathcal F_c}n_f=n_c\). We define the second-level relevance score as \(b_{if}=\bar q_i^{\top}\bar k_f/\sqrt d\). These finer summaries are used only to assess the reliability of the sub-block representatives.

\subsection{Query-Adaptive Sparse Routing}
\label{sec:qask_routing}

\paragraph{Candidate Standardized Relevance Threshold}
We calibrate the sparsity threshold separately for each query tile and each of its key sources. This allows different sources to receive different computation allocations for the same query, while the allocation within a source can also change across queries. To determine these thresholds, we evaluate the regions that remain approximated under each candidate route and check whether their estimated risk satisfies the budget of the corresponding source.  We first introduce \(\beta\), the candidate standardized relevance threshold. Within a key source, \(\beta\) controls which blocks are sparsified through the threshold \(t_i(\beta)=\mu_i+\beta\sigma_i\), defined below. We achieve adaptive routing by selecting \(\beta\). Each key source has multiple candidates. We use two quantities to select among them. The residual risk \(R_{i,g}(\beta)\) estimates the total approximation risk after sparsification. The risk budget \(\varepsilon_g\) is a hyperparameter that controls sparsity. We choose the largest candidate \(\beta\) that satisfies \(R_{i,g}(\beta)\leq\varepsilon_g\). In QASK-Edit, the residual risk need not vary monotonically with \(\beta\). For each candidate, we use \(\beta\) to determine the route and estimate the approximation risk associated with the resulting routing decisions. We evaluate candidates in descending order and stop at the first feasible candidate. We retain the routing rule of the selected candidate.

\paragraph{Approximation Risk}
We next describe how to estimate the approximation risk introduced by sparse computation. Although these quantities are evaluated as part of routing for a candidate \(\beta\), their definitions are independent of \(\beta\). They form an important part of \(R_{i,g}(\beta)\). LIVEditor observes a relationship between attention sharpness and pooling error~\cite{shao2026liveditor}. Inspired by this insight, we assess approximation reliability for each query tile and KV block. In our setting, a highly relevant but coherent block may be well represented by its mean. By contrast, a modest block score can hide a few strong local responses. We therefore jointly consider the estimated attention probability \(\hat p_{ij}\) of a block and its sub-block dispersion \(u_{ij}\):
\begin{equation}
\hat p_{ij}=\frac{n_j\exp(s_{ij})}{\sum_{k:n_k>0}n_k\exp(s_{ik})},
\label{eq:qask_mass}
\end{equation}
\begin{equation}
u_{ij}=\sum_{c\in\mathcal C_j} w_c(a_{ic}-s_{ij})^2.
\label{eq:qask_dispersion}
\end{equation}
The normalization in Eq.~\eqref{eq:qask_mass} includes all valid key sources in the current attention operation. Thus, \(\hat p_{ij}\) estimates a block's share of the joint attention mass. The factor \(n_j\) accounts for the number of tokens represented by its mean. In Eq.~\eqref{eq:qask_dispersion}, larger \(u_{ij}\) indicates greater disagreement among sub-block responses. We define the whole-block approximation risk as \(r_{ij}=\hat p_{ij}u_{ij}\). This assigns higher risk to blocks that receive more attention and are more heterogeneous.

\paragraph{Hierarchical Selection}
With \(r_{ij}\) defined, we return to the risk budget \(\varepsilon_g\). We use it both to screen reliable block summaries and to constrain a source's total residual risk. This shared choice is motivated by the probability weighting in \(r_{ij}=\hat p_{ij}u_{ij}\). Since \(\sum_{j:g(j)=g}\hat p_{ij}\leq 1\), any collection of its wholly approximated blocks with \(u_{ij}\leq\varepsilon_g\) contributes at most \(\varepsilon_g\) to the proxy risk. We therefore tie the local dispersion cutoff to the source-level budget rather than introduce a separate screening threshold. For clarity, the remaining approximate sub-blocks in partially refined blocks still contribute to the risk. We therefore also check the full \(R_{i,g}(\beta)\).

Let \(\mu_i\) and \(\sigma_i\) denote the token-count-weighted mean and standard deviation of block relevance scores. Video sources use all video blocks, while modalities requiring separate statistics use their own block groups. We omit the group index for clarity. For a candidate standardized relevance threshold \(\beta\), we define the relevance routing threshold as \(t_i(\beta)=\mu_i+\beta\sigma_i\). Thus, \(\beta\) specifies the cutoff in standard-deviation units, while \(t_i(\beta)\) expresses it on the relevance-score scale. Note that sources sharing the same statistics can still select different standardized thresholds.

For a candidate \(\beta\), let \(\mathcal A_{ij}(\beta)\subseteq\mathcal C_j\) denote the sub-blocks selected for exact, token-level computation. A protected sub-block is one explicitly designated for full computation. Any block containing a protected sub-block is computed entirely at token resolution. For such blocks, \(\mathcal A_{ij}(\beta)=\mathcal C_j\). For all other blocks, we determine this set in three cases. If \(u_{ij}\leq\varepsilon_{g(j)}\), we treat the block summary as reliable and approximate the entire block. Otherwise, if \(s_{ij}>t_i(\beta)\), we select all its valid sub-blocks. In this case, the block is heterogeneous and has a high relevance score. For the remaining blocks, we select only valid sub-blocks satisfying \(a_{ic}>t_i(\beta)\). This last case recovers localized responses that the block score would miss, while keeping the rest approximate.

\paragraph{Residual Risk}
After routing, we estimate the approximation risk associated with the regions that remain approximated. A wholly approximated block uses its block mean, so we estimate its approximation risk using \(u_{ij}\). In a partially refined block, each remaining sub-block uses its own mean. Its approximation risk therefore depends on variation within that sub-block. For \(f\in\mathcal F_c\), we define the second-level token fraction as \(\omega_f=n_f/n_c\). The within-sub-block dispersion \(u^{(2)}_{ic}\) is
\begin{equation}
u^{(2)}_{ic}=\sum_{f\in\mathcal F_c}\omega_f(a_{ic}-b_{if})^2.
\label{eq:qask_second_level_dispersion}
\end{equation}
This measures the disagreement between a sub-block representative and its two finer representatives. For clarity, second-level sub-blocks are used only for risk estimation. They introduce no additional decisions about exact computation.

The earlier quantity \(r_{ij}\) assumes that the entire block is represented by its block mean. We use the superscript \(\mathrm{res}\) to distinguish the residual risk after applying a candidate route. We use the block dispersion when no sub-block is selected. For a partially refined block, we use the within-sub-block dispersions of the remaining approximate sub-blocks. The residual risk is zero for fully computed blocks, including all blocks containing protected sub-blocks:
\begin{equation}
r^{\mathrm{res}}_{ij}(\beta)=\hat p_{ij}
\begin{cases}
u_{ij}, & \mathcal A_{ij}(\beta)=\varnothing,\\[2pt]
0, & \mathcal A_{ij}(\beta)=\mathcal C_j,\\[2pt]
\displaystyle\sum_{c\in\mathcal C_j\setminus\mathcal A_{ij}(\beta)}w_cu^{(2)}_{ic}, & \text{otherwise}.
\end{cases}
\label{eq:qask_block_residual}
\end{equation}
The total residual risk for key source \(g\) is
\begin{equation}
R_{i,g}(\beta)=\sum_{j:g(j)=g}r^{\mathrm{res}}_{ij}(\beta).
\label{eq:qask_residual}
\end{equation}
Because routing may switch between block and sub-block representatives, the residual risk need not vary monotonically with \(\beta\). In our experiments, we also evaluated an alternative selection rule. It checks all candidates and selects the feasible routing pattern with the lowest residual risk. Compared with accepting the first feasible candidate, this provided no clear improvement in editing quality. It also slightly increased end-to-end latency because of the extra routing overhead. We therefore use early stopping.

If no candidate is feasible, we compute every block with \(u_{ij}>\varepsilon_g\) at token resolution. Blocks containing protected sub-blocks are also computed entirely at token resolution. All remaining blocks use whole-block approximations. Sources with a zero budget or those designated for full computation always use their original tokens. Thus, \(\varepsilon_g\) controls source-specific routing through the residual-risk test. The threshold and exact-computation pattern adapt to the current query.

\subsection{Unified Exact and Approximate Attention}
\label{sec:qask_attention}

Once routing is finalized, each KV block contributes to the attention output in one of the following three ways:
\begin{itemize}
    \item \textbf{Whole-block approximation.} If no sub-block is selected, one pair of mean key and value vectors represents the entire block.
    \item \textbf{Partial refinement.} If only some sub-blocks are selected, we use their original KV tokens. Each remaining sub-block is represented by its mean key and value.
    \item \textbf{Full token-level computation.} If all valid sub-blocks are selected, the block contributes all its original KV tokens. Blocks containing protected sub-blocks always follow this case.
\end{itemize}

These contributions are combined across all key sources under a single softmax to obtain the final output.

\paragraph{Formal Definition}
We now formalize the above process. Let \(\mathcal T_i\) collect the resulting attention entries from all sources for query tile \(i\). Each entry \(b\) has a key \(\kappa_b\), a value \(\nu_b\), and a token multiplicity \(m_b\). An original KV token has \(m_b=1\). A block representative has \(m_b=n_j\), and a sub-block representative has \(m_b=n_c\). For each original query \(q_t\) in tile \(i\), let \(\ell_t\) collect its multiplicity-adjusted attention logits \(\ell_{tb}\). We compute these logits and the attention output \(o_t\) as
\begin{equation}
\begin{aligned}
\ell_{tb}&=\frac{q_t^{\top}\kappa_b}{\sqrt d}+\log m_b,\\
o_t&=\sum_{b\in\mathcal T_i}\operatorname{softmax}(\ell_t)_b\,\nu_b,
\end{aligned}
\label{eq:qask_output}
\end{equation}
where the softmax is taken over all entries in \(\mathcal T_i\). The \(\log m_b\) term accounts for the number of tokens represented by each pooled entry. The same aggregation rule therefore handles all three cases.

\section{Experiment}
\subsection{Experimental Setups}
\label{sec:experimental_setups}

\begin{table*}[t]
\centering
\begingroup
\small
\setlength{\tabcolsep}{2pt}
\renewcommand{\arraystretch}{1.10}
\definecolor{qaskrowgray}{gray}{0.92}
\newcommand{\qaskgain}[2]{\mbox{#1\hspace{0.12em}\raisebox{-0.30ex}{{\normalfont\footnotesize #2}}}}
\newcommand{\qaskbest}[1]{\textbf{#1}}
\newcommand{\qasksecond}[1]{\underline{#1}}
\newlength{\qaskmethodwidth}
\newlength{\qaskmetricwidth}
\newlength{\qaskssimwidth}
\setlength{\qaskmethodwidth}{0.145\linewidth}
\setlength{\qaskmetricwidth}{\dimexpr(\linewidth-\qaskmethodwidth-14\tabcolsep-6pt)/7\relax}
\setlength{\qaskssimwidth}{\dimexpr\qaskmetricwidth+6pt\relax}
\caption{Quantitative comparison of video editing quality and efficiency across three video editing models.}
\label{tab:qask-main-results-v2-2}
\begin{tabular}{@{}>{\bfseries\raggedright\arraybackslash}p{\qaskmethodwidth}*{2}{>{\centering\arraybackslash}p{\qaskmetricwidth}}>{\centering\arraybackslash}p{\qaskssimwidth}*{4}{>{\centering\arraybackslash}p{\qaskmetricwidth}}@{}}
\toprule
\multirow{2}{*}{\textbf{Method}}
& \multicolumn{4}{c}{\textbf{FiVE-Bench}}
& \multicolumn{2}{c}{\textbf{IVE-Bench}}
& \textbf{Efficiency} \\
\cmidrule(lr){2-5}\cmidrule(lr){6-7}\cmidrule(l){8-8}
& \textbf{FiVE-Acc}$\uparrow$ & \textbf{PSNR}$\uparrow$
& \textbf{SSIM}$\uparrow$ & \textbf{LPIPS}$\downarrow$
& \textbf{IVE-short}$\uparrow$ & \textbf{IVE-long}$\uparrow$
& \textbf{Speedup}$\uparrow$ \\
\midrule
\multicolumn{8}{c}{\textbf{\textit{StreamEdit}}} \\
\midrule
Dense & 0.5826 & 23.42 & 0.7446 & 0.1199 & 0.6741 & 0.6840 & 1.00$\times$ \\
SpargeAttention & 0.5725 & 23.38 & 0.7433 & 0.1205 & 0.6675 & 0.6864 & \qasksecond{1.31}$\times$ \\
SVG2 & 0.5267 & 23.31 & 0.7407 & 0.1222 & 0.6603 & 0.6832 & 1.12$\times$ \\
SVOO & 0.5593 & 23.36 & 0.7425 & 0.1210 & 0.6676 & 0.6828 & {1.29}$\times$ \\
Sol-Attn & 0.5743 & 23.42 & 0.7445 & 0.1203 & 0.6708 & 0.6856 & \qasksecond{1.31}$\times$ \\
ISA & 0.5987 & 21.00 & 0.7097 & 0.1520 & 0.6912 & 0.6967 & 1.03$\times$ \\
\rowcolor{qaskrowgray}
QASK-Edit {\normalfont (Ours)} & \qaskgain{0.5819}{-0.13\%} & \qaskgain{23.20}{-0.92\%} & \qaskgain{0.7409}{-0.49\%} & \qaskgain{0.1237}{-3.11\%} & \qaskgain{0.6613}{-1.90\%} & \qaskgain{0.6861}{+0.32\%} & \qaskgain{\qaskbest{1.46}$\times$}{+46.02\%} \\
\midrule
\multicolumn{8}{c}{\textbf{\textit{UniEdit-Flow}}} \\
\midrule
Dense & 0.2028 & 25.16 & 0.7526 & 0.1465 & 0.6427 & 0.6515 & 1.00$\times$ \\
SpargeAttention & 0.1802 & 24.29 & 0.7378 & 0.1594 & 0.6461 & 0.6596 & \qasksecond{1.21}$\times$ \\
SVG2 & 0.2053 & 25.12 & 0.7519 & 0.1473 & 0.6455 & 0.6635 & 1.10$\times$ \\
SVOO & 0.2036 & 25.13 & 0.7519 & 0.1472 & 0.6393 & 0.6596 & 1.14$\times$ \\
Sol-Attn & 0.1894 & 24.68 & 0.7447 & 0.1522 & 0.6385 & 0.6666 & 1.11$\times$ \\
ISA & 0.2689 & 23.16 & 0.7110 & 0.1810 & 0.6534 & 0.6517 & 1.07$\times$ \\
\rowcolor{qaskrowgray}
QASK-Edit {\normalfont (Ours)} & \qaskgain{0.2470}{+21.83\%} & \qaskgain{24.06}{-4.37\%} & \qaskgain{0.7355}{-2.27\%} & \qaskgain{0.1534}{-4.72\%} & \qaskgain{0.6444}{+0.27\%} & \qaskgain{0.6548}{+0.50\%} & \qaskgain{\qaskbest{1.26}$\times$}{+26.44\%} \\
\midrule
\multicolumn{8}{c}{\textbf{\textit{MiniMax-H3}}} \\
\midrule
Dense & 0.4879 & 12.92 & 0.5017 & 0.4496 & 0.5112 & 0.5383 & 1.00$\times$ \\
SpargeAttention & 0.5104 & 12.96 & 0.5021 & 0.4461 & 0.5477 & 0.5761 & \qasksecond{1.31}$\times$ \\
SVG2 & 0.4848 & 12.99 & 0.4993 & 0.4499 & 0.5196 & 0.5540 & 1.08$\times$ \\
SVOO & 0.4788 & 13.15 & 0.5053 & 0.4373 & 0.6035 & 0.6241 & $1.19\times$ \\
Sol-Attn & 0.5038 & 12.77 & 0.4996 & 0.4542 & 0.5776 & 0.5932 & 1.09$\times$ \\
ISA & 0.4504 & 12.81 & 0.4949 & 0.4522 & 0.5754 & 0.5813 & 1.29$\times$ \\
\rowcolor{qaskrowgray}
QASK-Edit {\normalfont (Ours)} & \qaskgain{0.4572}{-6.28\%} & \qaskgain{12.92}{+0.03\%} & \qaskgain{0.5031}{+0.27\%} & \qaskgain{0.4482}{+0.32\%} & \qaskgain{0.6076}{+18.86\%} & \qaskgain{0.6236}{+15.85\%} & \qaskgain{\qaskbest{1.33}$\times$}{+33.45\%} \\
\bottomrule
\end{tabular}
\par\vspace{4pt}
\begin{minipage}{\linewidth}
\footnotesize
\end{minipage}
\endgroup
\vspace{-1em}
\end{table*}

\paragraph{Models}
We evaluate {QASK-Edit} on three video editors: StreamEdit~\cite{jiao2026streamedit}, UniEdit-Flow~\cite{jiao2026unieditflow}, and MiniMax-H3~\cite{minimax2026h3}. These cover inversion-free streaming generation, inversion-based flow editing, and pretrained multimodal instruction-guided editing, respectively. We use the Self Forcing variant of StreamEdit and the official Wan2.1-based implementation of UniEdit-Flow.

\paragraph{Baselines}
We compare against dense attention and five training-free sparse attention methods: SpargeAttention~\cite{zhang2025spargeattention}, SVG2~\cite{yang2025svg2}, SVOO~\cite{luo2026svoo}, Sol-Attn~\cite{li2026solattn}, and In-context Sparse Attention (ISA) from LIVEditor~\cite{shao2026liveditor}. We follow the hyperparameter configurations provided in their official repositories.

\paragraph{Benchmarks and metrics}
We evaluate on FiVE-Bench~\cite{li2025five} and IVEBench~\cite{chen2026ivebench}. On FiVE-Bench, we report the VLM-based FiVE-Acc to assess editing success and use PSNR, LPIPS~\cite{zhang2018lpips}, and SSIM~\cite{wang2004ssim} to evaluate unedited-region fidelity. On IVEBench, we report the official Total Score separately on the Short and Long subsets. We measure per-video end-to-end editing latency and report speedups relative to each backbone’s released dense-attention pipeline, excluding one-time model loading, compilation, and quantization.

\paragraph{Implementation details}
We use 64-token query tiles and 128-token KV blocks, each divided into four 32-token sub-blocks. The candidate standardized relevance thresholds are \(\mathcal B=\{-1,0,0.5,1,1.5,2,3,4\}\). For fairness, resolutions and frame counts follow the respective open-source implementations, adjusted for benchmark video lengths (see Appendix~\ref{app:comparison_protocol}). Comparisons use fair settings within each editor for sparse methods. More details are provided in Appendix~\ref{app:experimental_details}. All experiments are conducted on NVIDIA RTX 5880 Ada GPUs with 48\,GB of memory.

\subsection{Main Results}
\label{sec:main_results}

Table~\ref{tab:qask-main-results-v2-2} summarizes the main results across the three video editors. {QASK-Edit} achieves \(1.46\times\), \(1.26\times\), and \(1.33\times\) end-to-end speedups on StreamEdit, UniEdit-Flow, and MiniMax-H3, respectively, yielding the highest speedup among the reported methods for each editor.

The quality--efficiency trade-off varies across editors. On StreamEdit, FiVE-Acc remains close to dense attention, with a \(1.90\%\) relative decrease in the IVE Short score. On UniEdit-Flow, FiVE-Acc improves by \(21.83\%\) relative to dense attention, while PSNR decreases by \(1.10\)\,dB and both IVE scores remain close to dense. Compared with ISA on these two editors, {QASK-Edit} provides higher unedited-region fidelity and faster execution, although ISA attains higher FiVE-Acc.

On MiniMax-H3, {QASK-Edit} achieves the highest IVE Short score and an IVE Long score close to the best baseline, improving over dense attention by \(18.86\%\) and \(15.85\%\), respectively. Unedited-region fidelity remains close to dense, but FiVE-Acc decreases by \(6.28\%\). These results demonstrate consistent end-to-end acceleration across the evaluated editors, with model- and benchmark-dependent trade-offs in editing quality.

\subsection{Ablation Study}
\label{sec:ablation_study}

\paragraph{Source-specific and query-adaptive routing}
We ablate source-specific and query-adaptive threshold selection (Table~\ref{tab:ablation-sq}). To evaluate source-specific calibration, Shared-\(\beta\) selects a shared standardized threshold for each query tile that satisfies the risk budgets of all sources. QASK-Edit allows each source to select its threshold independently within its budget. Table\ref{tab:ablation-sq}(a) shows that source-specific calibration yields higher speedups on all three editors, while quality changes depend on the editor and evaluation metric. To evaluate online threshold calibration within each source, we calibrate a fixed \(\beta\) for each source on 24 samples excluded from evaluation, matching its average exact-computation ratio to that of {QASK-Edit}. Table~\ref{tab:ablation-sq}(b) shows modest improvements in both IVE on all three editors, while FiVE-Acc changes are mixed.

\begin{table}[tbp]
\centering
\caption{Ablation of source-specific and query-adaptive routing.}
\label{tab:ablation-sq}

\begingroup
\fontsize{8}{9.5}\selectfont
\setlength{\tabcolsep}{2.6pt}
\renewcommand{\arraystretch}{1.12}

\noindent\textbf{(a) Source-specific routing}\par\vspace{2.5pt}
\begin{tabular*}{\linewidth}{@{\extracolsep{\fill}}llrrrr@{}}
\toprule
Model & Routing & Speed\(\uparrow\) & FiVE\(\uparrow\) & IVE-S\(\uparrow\) & IVE-L\(\uparrow\) \\
\midrule
\multirow{2}{*}{StreamEdit} & Shared-\(\beta\) & 1.30\(\times\) & \textbf{58.31} & \textbf{67.62} & 68.49 \\
 & Full & \textbf{1.46\(\times\)} & 58.19 & 66.13 & \textbf{68.61} \\
\addlinespace[3pt]
\multirow{2}{*}{UniEdit-Flow} & Shared-\(\beta\) & 1.19\(\times\) & \textbf{24.70} & \textbf{64.44} & \textbf{65.59} \\
 & Full & \textbf{1.26\(\times\)} & \textbf{24.70} & \textbf{64.44} & 65.48 \\
\addlinespace[3pt]
\multirow{2}{*}{MiniMax-H3} & Shared-\(\beta\) & 1.29\(\times\) & 41.54 & 60.13 & 61.24 \\
 & Full & \textbf{1.33\(\times\)} & \textbf{45.72} & \textbf{60.76} & \textbf{62.36} \\
\bottomrule
\end{tabular*}

\par\vspace{7pt}

\noindent\textbf{(b) Online vs. calibrated source-fixed \(\beta\)}\par\vspace{2.5pt}
\begin{tabular*}{\linewidth}{@{\extracolsep{\fill}}lrrr@{}}
\toprule
Model & \(\Delta\)FiVE & \(\Delta\)IVE-S & \(\Delta\)IVE-L \\
\midrule
StreamEdit & \textbf{+1.75} & \textbf{+0.26} & \textbf{+0.03} \\
UniEdit-Flow & -0.11 & \textbf{+0.62} & \textbf{+0.70} \\
MiniMax-H3 & -0.80 & \textbf{+0.50} & \textbf{+1.18} \\
\bottomrule
\end{tabular*}
\par\vspace{3pt}
\endgroup
\vspace{-0.5em}
\end{table}

\begin{figure}[tbp]
\centering
\includegraphics[width=\linewidth]{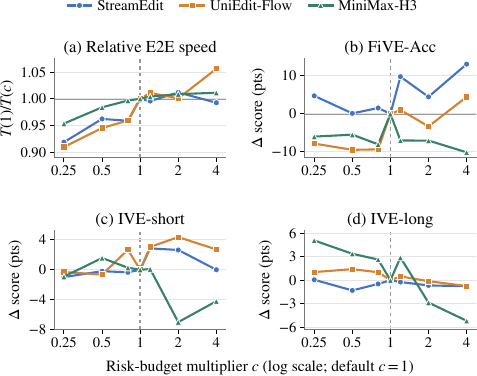}
\caption{\textbf{Sensitivity to the risk budget.} Default budgets are scaled by \(c\), and \(T(c)\) denotes generation latency. Quality changes are measured in score points relative to \(c=1\).}
\label{fig:ablation_epilson}
\vspace{-1em}
\end{figure}

\paragraph{Sensitivity to the risk budget}
We scale each editor's default risk budgets as \(\varepsilon_g=c\,\varepsilon_g^{\mathrm{default}}\) on a topic-stratified 200-video evaluation subset, with sampling details provided in the appendix. Figure~\ref{fig:ablation_epilson} shows that tighter budgets generally slow inference, whereas larger budgets yield limited additional acceleration. Quality responses depend on the editor and metric: IVE-Long exhibits a clearer overall trade-off, especially on MiniMax-H3, while the other scores vary nonmonotonically. Additional pilot tests suggest further slowdown with tighter budgets and diminishing speed gains from further relaxation. Thus, the risk budget provides an efficiency control with task-dependent quality responses.

\begin{table}[tbp]
\centering
\caption{Relative reductions compared with the offline-calibrated source-fixed \(\beta\) baseline. Each entry reports attention output error reduction / exact block ratio reduction (\%). Negative exact block ratio reductions indicate increased allocation. \(o\) and \(e\) denote original and edited streams.}
\label{tab:qask_attention_error_analysis}
\renewcommand{\arraystretch}{1.2} %
\footnotesize
\begin{tabular*}{\linewidth}{@{\extracolsep{\fill}}lccc@{\extracolsep{\fill}}}
\toprule
\(Q \rightarrow K\) & Low & Mid & High \\
\midrule
\(Q_o \rightarrow K_o\) & 13.9\% / 2.4\% & 15.2\% / 0.3\% & 17.3\% / -2.5\% \\
\(Q_e \rightarrow K_e\) & 11.3\% / 2.3\% & 13.8\% / 0.1\% & 15.7\% / -2.0\% \\
\(Q_e \rightarrow K_o\) & 3.2\% / 6.5\% & 15.7\% / -2.0\% & 15.6\% / -6.2\% \\
\bottomrule
\end{tabular*}
\vspace{-0.8em}
\end{table}

\paragraph{Adaptation to attention demand}
To examine adaptation to attention demand, we compare {QASK-Edit} with the calibrated source-fixed baseline on the same queries. We group queries by their required block ratios into low, medium, and high demand, corresponding to the bottom \(25\%\), middle \(50\%\), and top \(25\%\), respectively, and measure attention output error against dense attention. Across all nine source, demand groups in Table~\ref{tab:qask_attention_error_analysis}, {QASK-Edit} reduces attention output error by 3.2\%--17.3\% relative to the source-fixed baseline. These results show that online calibration adapts computation allocation to query demand, reducing computation in low-demand groups while allocating more to high-demand groups.

\subsection{Generalization to Reference Conditioned Generation}
\label{sec:generalization}

\begin{table}[ht]
  \centering
  \caption{Generalization to reference-conditioned video generation.}
  \label{tab:generalization-single}
  \begingroup

  \providecommand{\GenBackboneA}{Wan-AI/Wan2.1-VACE-1.3B}
  \providecommand{\GenBackboneB}{bytedance-research/Phantom}

  \footnotesize
  \setlength{\tabcolsep}{2pt}
  \renewcommand{\arraystretch}{1.12}

  \begin{tabularx}{\columnwidth}{l*{3}{>{\centering\arraybackslash}X}}
    \toprule
    Method & Speedup $\uparrow$
           & OpenS2V-Eval $\uparrow$
           & VACE-R2V $\uparrow$ \\
    \midrule

    \multicolumn{4}{@{}l}{\textit{\GenBackboneA}} \\[1pt]

    Dense       & 1.00$\times$ & 45.51 & 75.06 \\
    SVG2        & 1.22$\times$ & 45.14 & 74.73 \\
    SpargeAttn  & 1.44$\times$ & 43.46 & 75.22 \\
    Sol-Attn    & 1.32$\times$ & 45.18 & 75.10 \\

    \addlinespace[2pt]
    \rowcolor{black!9}
    \textbf{QASK-Edit}
      & \qaskdelta{{1.36$\times$}}{+36.00\%}
      & \qaskdelta{46.52}{+2.22\%}
      & \qaskdelta{75.17}{+0.13\%} \\

    \midrule

    \multicolumn{4}{@{}l}{\textit{\GenBackboneB}} \\[1pt]

    Dense       & 1.00$\times$ & 46.80 & 74.83 \\
    SVG2        & 1.22$\times$ & 47.35 & 74.54 \\
    SpargeAttn  & 1.45$\times$ & 45.66 & 75.58 \\
    Sol-Attn    & 1.29$\times$ & 46.51 & 74.79 \\

    \addlinespace[2pt]
    \rowcolor{black!9}
    \textbf{QASK-Edit}
      & \qaskdelta{{1.32$\times$}}{+32.00\%}
      & \qaskdelta{48.02}{+2.61\%}
      & \qaskdelta{74.84}{+0.01\%} \\

    \bottomrule
  \end{tabularx}

  \endgroup
\vspace{-0.5em}
\end{table}

To assess generalization beyond video editing, we evaluate {QASK-Edit} on reference-conditioned video generation with Wan2.1-VACE-1.3B~\cite{jiang2025vace} and Phantom~\cite{liu2025phantom}. We report the open-domain Total Score on OpenS2V-Eval~\cite{yuan2025opens2v} and the Normalized Average of VBench~\cite{huang2024vbench} metrics on the R2V subset of VACE-Benchmark, following the aggregation used in VACE. As shown in Table~\ref{tab:generalization-single}, our method achieves \(1.36\times\) and \(1.32\times\) speedups, respectively. Relative to dense attention, OpenS2V-Eval scores increase by \(1.01\) and \(1.22\) points, while VACE-R2V scores remain nearly unchanged. These results suggest that query-adaptive computation allocation also enables efficient reference-conditioned generation while maintaining quality on the evaluated benchmarks.

\section{Conclusion}

In this work, we revisit sparse attention for generation-based video editing from the perspective of heterogeneous key sources. We observe that attention sparsity varies substantially across key sources and individual queries, making fixed sparsity allocation inefficient. To address this challenge, we introduce \textbf{QASK-Edit}, a training-free sparse attention framework that adaptively distributes computation across key sources and queries according to the estimated approximation risk. Extensive experiments across three diverse video editing models demonstrate up to 1.46× end-to-end speedups while maintaining competitive editing quality. Furthermore, evaluations on reference-conditioned video generation demonstrate the applicability of our approach beyond video editing. These findings highlight the importance of source-aware, query-adaptive computation allocation for efficiently accelerating video generation.

\clearpage
\section*{LLM Usage Statement}

We used ChatGPT (OpenAI) and Claude (Anthropic) as assistive tools during this research. For manuscript preparation, ChatGPT is used for grammar checking, verifying the logical flow of linguistic statements, and translating certain texts, with all translated text subsequently revised and finalized by the authors. During the research, ChatGPT and Claude were also used to assess the rigor of experimental designs and evaluation procedures. For figures and tables, Claude was used to review layout. For implementation, ChatGPT was primarily used for code review, while LLMs were used to generate scripts for repetitive experimental tasks. All research decisions, experimental results, and final manuscript content remain the responsibility of the authors.

{
    \small
    \bibliographystyle{ieeenat_fullname}
    \bibliography{main}
}

\clearpage
\setcounter{page}{1}
\maketitlesupplementary
\appendix
\renewcommand{\thepage}{S\arabic{page}}
\setcounter{table}{0}
\renewcommand{\thetable}{S\arabic{table}}
\renewcommand{\theHtable}{supp.\arabic{table}}
\makeatletter
\setlength{\@dblfptop}{0pt}
\setlength{\@dblfpsep}{14pt}
\setlength{\@dblfpbot}{0pt plus 1fil}
\makeatother

\section{Implementation Details}
\label{app:experimental_details}

\subsection{Comparison}
\label{app:comparison_protocol}

All comparison methods use common RoPE and query/key RMS normalization fusions, together with author-provided optimized operators where available. Reported speedups measure the end-to-end performance of these implementations relative to Dense.

As in Sec.~\ref{sec:experimental_setups}, we measure per-video end-to-end generation latency, excluding one-time model loading, compilation, and quantization. All methods within a backbone use the same hardware placement and generation settings.

\subsection{Backbone Configurations}
\label{app:backbone_configurations}

Resource constraints require adjustments to some backbone defaults. We apply each adjustment consistently across Dense and all sparse methods evaluated on that backbone.

\paragraph{StreamEdit}
We use the Self Forcing variant of StreamEdit~\cite{jiao2026streamedit} with the official spatial resolution of \(832\times480\) and a frame rate of \(16\) fps. The total frame count is adapted to the duration of each benchmark video, with a maximum of \(81\) frames.

\paragraph{UniEdit-Flow}
For the Wan2.1-based UniEdit-Flow implementation~\cite{jiao2026unieditflow}, we set the maximum pixel area to \(832\times480\). Inputs exceeding this area are first resized isotropically to preserve their aspect ratio. The resulting width and height are then rounded down independently to multiples of \(16\), as required by the Wan pipeline. Output resolutions can therefore differ across videos; \(832\times480\) is the most common resolution in our evaluation.

\paragraph{MiniMax-H3}
We evaluate MiniMax-H3~\cite{minimax2026h3} at \(256\times256\) with a fixed length of \(124\) frames to accommodate the available compute resources. We start from the official BF16 checkpoint and use \texttt{torchao}~\cite{torchao2024} to quantize the model to INT8 for inference. The reference transformer, named \texttt{transformer\_ref} in the implementation, occupies a dedicated GPU. The Qwen encoder occupies another dedicated GPU, while the video and audio VAEs share a further GPU. Thus, this configuration uses component placement across multiple GPUs. The same quantized model and component placement are used for all compared attention methods.

\subsection{QASK-Edit Implementation}
\label{app:qask_implementation}

\paragraph{Fused kernels}
We implement the online summarization, routing, and attention operators in Triton~\cite{tillet2019triton} to reduce the overhead of adaptive sparsification. First, KV reordering and packing are performed together with the computation of block, sub-block, and second-level sub-block means. Second, risk estimation, candidate \(\beta\) selection, mask construction, and exact-computation index generation are fused. Third, the token-multiplicity correction is incorporated directly into the softmax, and the token-level and representative-based paths are evaluated within a fused attention implementation. All selected tokens and representatives contribute to the same softmax normalization.

\paragraph{Hierarchical layouts}
We use \(64\)-token query tiles and \(128\)-token KV blocks. The candidate KV block layouts in \((T,H,W)\) order are \(1\times8\times16\), \(8\times1\times16\), and \(4\times2\times16\), corresponding to spatial, temporal, and mixed layouts, respectively. Each complete KV block contains four \(32\)-token sub-blocks, and each sub-block contains two \(16\)-token second-level sub-blocks. Following the SICS-based layout selection in Sec.~\ref{sec:qask_blocks}, queries and video key sources choose their layouts independently. The child layouts follow the selected parent layout, as summarized in Table~\ref{tab:suppl_layouts}. Token counts refer to valid tokens when a boundary block is incomplete. Second-level representatives are used only for risk estimation (already mentioned in the main text).

\begin{table}[t]
  \centering
  \caption{Hierarchical KV layouts in \((T,H,W)\) order. Each complete block, sub-block, and second-level sub-block contains \(128\), \(32\), and \(16\) tokens, respectively.}
  \label{tab:suppl_layouts}
  \begingroup
  \footnotesize
  \setlength{\tabcolsep}{3pt}
  \renewcommand{\arraystretch}{1.15}
  \begin{tabularx}{\columnwidth}{@{}l*{3}{>{\centering\arraybackslash}X}@{}}
    \toprule
    Layout & Block & Sub-block & Second level \\
    \midrule
    Spatial & \(1\times8\times16\) & \(1\times2\times16\) & \(1\times1\times16\) \\
    Temporal & \(8\times1\times16\) & \(2\times1\times16\) & \(1\times1\times16\) \\
    Mixed & \(4\times2\times16\) & \(1\times2\times16\) & \(1\times1\times16\) \\
    \bottomrule
  \end{tabularx}
  \endgroup
\end{table}

\paragraph{Routing configuration}
The candidate standardized relevance thresholds are \(\mathcal B=\{-1,0,0.5,1,1.5,2,3,4\}\). We evaluate them in descending order and accept the first feasible candidate according to the residual-risk test in Sec.~\ref{sec:qask_routing}. The source-specific risk budget \(\varepsilon_g\) is the sparsity-control hyperparameter; its default values for the video editors are given in Table~\ref{tab:suppl_budgets}. Here, reference video denotes the conditioning source video, and generated video denotes the video being edited. Risk-budget ablations multiply all default budgets of an editor by the same factor \(c\), following \(\varepsilon_g=c\,\varepsilon_g^{\mathrm{default}}\).

\begin{table}[t]
  \centering
  \caption{Default source-specific risk budgets \(\varepsilon_g\) for the editing experiments. A dash indicates that no audio-source budget is used in the corresponding configuration.}
  \label{tab:suppl_budgets}
  \begingroup
  \footnotesize
  \setlength{\tabcolsep}{3pt}
  \renewcommand{\arraystretch}{1.15}
  \begin{tabularx}{\columnwidth}{@{}l*{3}{>{\centering\arraybackslash}X}@{}}
    \toprule
    Backbone & Reference video & Generated video & Audio \\
    \midrule
    StreamEdit & 0.10 & 0.10 & --- \\
    UniEdit-Flow & 0.02 & 0.02 & --- \\
    MiniMax-H3 & 0.30 & 0.10 & 0.10 \\
    \bottomrule
  \end{tabularx}
  \endgroup
\end{table}

\section{Design Rationale for Query–Source Adaptive Allocation}

The design of QASK-Edit follows from heterogeneous attention requirements across key sources and across queries within each source. These two dimensions motivate calibrating sparsity separately for each query tile and key source. A threshold shared across sources couples their allocations through a common standardized cutoff, while a fixed cutoff for each source captures persistent source differences but does not adapt the cutoff to the current query. QASK-Edit permits both forms of adaptation by selecting a source-specific cutoff for each query tile.

Source-specific risk budgets specify tolerances on estimated approximation risk rather than fixed retention ratios. A fixed budget can therefore lead to different thresholds and exact-computation patterns for different queries. Likewise, sharing relevance-score statistics across video sources provides a common reference scale without requiring the same standardized cutoff. A fixed cutoff can itself produce input-dependent routing patterns; the additional adaptation in QASK-Edit is the online selection of that cutoff for each query tile and source.

This allocation requires a criterion for deciding how much exact computation to retain for each query–source pair. We use estimated attention mass together with sub-block score dispersion to assess the contributions that remain approximated under a candidate route. The resulting risk estimate provides a criterion for selecting a threshold within each source’s budget. It is a proxy for approximation difficulty, rather than a certified bound on attention output error.

Hierarchical routing provides finer control over how the selected allocation is realized. A block can be represented by its mean, refined selectively at the sub-block level, or computed entirely at token resolution. This allows exact computation to target local interactions within a source. Table~\ref{tab:ablation-sq} examines independent calibration across sources and online calibration within each source, while Table~\ref{tab:qask_attention_error_analysis} examines how the resulting computation allocation changes with query demand.

\section{Evaluation Details}
\label{app:evaluation_details}

\subsection{Evaluation Scope and Random Seeds}
\label{app:evaluation_scope}

Unless explicitly stated otherwise in the main paper, the main comparisons, ablations, and generalization experiments use the benchmark sets or subsets named for the corresponding experiment, with calibration samples excluded from evaluation. The risk-budget sensitivity study uses a stratified subset of \(200\) editing tasks, with the sampling procedure detailed below.

We set the generation seed to \(2026\) for editing comparisons and to \(42\) for generalizaion experiments (section~\ref{sec:generalization}). Within each experiment, the same seed setting is used for all compared methods.

\paragraph{Source-fixed calibration}
We select \(24\) calibration samples using a sampling seed of \(2026\). FiVE-Bench samples are drawn by stratified sampling across categories \(1\)--\(6\), while IVEBench Short and Long samples are drawn randomly within their respective subsets. All calibration samples are excluded from evaluation, yielding disjoint calibration and evaluation sets. We calibrate one fixed \(\beta\) per source by matching the average exact-computation ratio of {QASK-Edit} on these samples.

\subsection{Stratified Sampling for Risk-Budget Ablations}
\label{app:stratified_sampling}

We draw \(200\) editing tasks from a pool of \(1{,}019\) tasks: \(419\) from FiVE-Bench, \(400\) from IVEBench Short, and \(200\) from IVEBench Long. Table~\ref{tab:suppl_sampling} summarizes the benchmark-level allocation.

\begin{table}[t]
  \centering
  \caption{Benchmark-level allocation of the \(200\)-task risk-budget ablation subset. Proportional quotas are shown to three decimals; allocation uses their unrounded values.}
  \label{tab:suppl_sampling}
  \begingroup
  \footnotesize
  \setlength{\tabcolsep}{3pt}
  \renewcommand{\arraystretch}{1.15}
  \begin{tabularx}{\columnwidth}{@{}l*{3}{>{\centering\arraybackslash}X}@{}}
    \toprule
    Benchmark & Total tasks & Proportional quota & Selected \\
    \midrule
    FiVE-Bench & 419 & 82.237 & 82 \\
    IVE Short & 400 & 78.508 & 79 \\
    IVE Long & 200 & 39.254 & 39 \\
    \midrule
    Total & 1019 & 200 & 200 \\
    \bottomrule
  \end{tabularx}
  \endgroup
\end{table}

Quotas are allocated hierarchically across benchmarks, categories, and subcategories. At each level, we apply the largest remainder method: proportional quotas are first rounded down, and the remaining slots are assigned in descending order of their fractional remainders. Ties are resolved by the lexicographic order of category labels.

Within each finest-grained subcategory, we sample tasks randomly using a base sampling seed of \(2026\). Each subcategory has an independent seed derived from the base seed and its benchmark, category, and subcategory identifiers. Subcategories with fewer available samples are processed first, and tasks whose source videos have already been selected are excluded. This ordering prioritizes scarce subcategories, while the exclusion rule avoids selecting multiple editing instructions for the same source video.

After sampling, we shuffle the execution order once using a fixed random seed. All seven risk-budget settings and all three editing models share the same sample list and the same execution order within each benchmark, without resampling. The sampling seed of \(2026\) is separate from the generation seed of \(42\) used for these ablations.

\subsection{Evaluator Adaptations}
\label{app:evaluator_adaptations}

We make two substitutions in the benchmark evaluators. For the API-dependent components of OpenS2V-Eval~\cite{yuan2025opens2v}, we use DeepSeek-V4.1-Flash~\cite{deepseek2026v41flash} in place of GPT-4o. For the VLM-based evaluation in FiVE-Bench~\cite{li2025five}, we use Qwen3-VL~\cite{bai2025qwen3vl} in place of Qwen2.5-VL. Each substitution is applied consistently to Dense and all sparse methods within the corresponding evaluation.

\begin{table*}[t]
  \centering
  \caption{Aggregate generalization scores and overall end-to-end speedups relative to Dense. Quality scores use a \(0\)--\(100\) scale; the VACE-R2V column retains additional precision from the normalized-score output.}
  \label{tab:suppl_gen_scores}
  \begingroup
  \footnotesize
  \setlength{\tabcolsep}{3pt}
  \renewcommand{\arraystretch}{1.16}
  \begin{tabularx}{\textwidth}{@{}l*{5}{>{\raggedleft\arraybackslash}X}@{}}
    \toprule
    Method & Speedup & \shortstack{OpenS2V\\Open} & \shortstack{OpenS2V\\Single} & \shortstack{OpenS2V\\Human} & \shortstack{VACE-R2V\\Norm. Avg.} \\
    \midrule
    \multicolumn{6}{@{}l}{\textit{Wan2.1-VACE-1.3B}} \\[2pt]
    Dense & 1.00\(\times\) & 45.51 & 49.1354 & 49.7747 & 75.0641 \\
    SVG2 & 1.22\(\times\) & 45.14 & 49.7207 & 49.3312 & 74.7346 \\
    SpargeAttn & 1.44\(\times\) & 43.46 & 44.7765 & 46.6589 & 75.2248 \\
    Sol-Attn & 1.32\(\times\) & 45.18 & 47.6974 & 46.9560 & 75.0960 \\
    \rowcolor{black!9}
    \textbf{QASK-Edit} & 1.36\(\times\) & 46.52 & 49.1497 & 50.1084 & 75.1665 \\
    \midrule
    \multicolumn{6}{@{}l}{\textit{Phantom}} \\[2pt]
    Dense & 1.00\(\times\) & 46.80 & 47.7525 & 51.0792 & 74.8262 \\
    SVG2 & 1.22\(\times\) & 47.35 & 49.1764 & 51.5934 & 74.5350 \\
    SpargeAttn & 1.45\(\times\) & 45.66 & 46.3112 & 49.7007 & 75.5765 \\
    Sol-Attn & 1.29\(\times\) & 46.51 & 47.0664 & 51.0965 & 74.7899 \\
    \rowcolor{black!9}
    \textbf{QASK-Edit} & 1.32\(\times\) & 48.02 & 49.1804 & 53.1950 & 74.8444 \\
    \bottomrule
  \end{tabularx}
  \endgroup
\end{table*}

\begin{table*}[t]
  \centering
  \caption{Raw VBench metrics on OpenS2V-Eval, rounded to four decimals without percentage scaling. Subject and background denote subject consistency and background consistency, respectively.}
  \label{tab:suppl_opens2v_vbench}
  \begingroup
  \footnotesize
  \setlength{\tabcolsep}{3pt}
  \renewcommand{\arraystretch}{1.16}
  \begin{tabularx}{\textwidth}{@{}l*{6}{>{\raggedleft\arraybackslash}X}@{}}
    \toprule
    Method & \shortstack{Subject\\consistency} & \shortstack{Background\\consistency} & \shortstack{Aesthetic\\quality} & \shortstack{Imaging\\quality} & \shortstack{Motion\\smoothness} & \shortstack{Dynamic\\degree} \\
    \midrule
    \multicolumn{7}{@{}l}{\textit{Wan2.1-VACE-1.3B}} \\[2pt]
    Dense & 0.9657 & 0.9581 & 0.5963 & 0.7004 & 0.9908 & 0.3152 \\
    SVG2 & 0.9620 & 0.9514 & 0.5920 & 0.6966 & 0.9902 & 0.3061 \\
    SpargeAttn & 0.9619 & 0.9554 & 0.5885 & 0.6853 & 0.9904 & 0.3303 \\
    Sol-Attn & 0.9640 & 0.9564 & 0.5924 & 0.6973 & 0.9906 & 0.3424 \\
    \rowcolor{black!9}
    \textbf{QASK-Edit} & 0.9637 & 0.9536 & 0.5936 & 0.6990 & 0.9908 & 0.3242 \\
    \midrule
    \multicolumn{7}{@{}l}{\textit{Phantom}} \\[2pt]
    Dense & 0.9665 & 0.9564 & 0.5917 & 0.7070 & 0.9899 & 0.4455 \\
    SVG2 & 0.9624 & 0.9502 & 0.5821 & 0.7028 & 0.9892 & 0.4182 \\
    SpargeAttn & 0.9636 & 0.9545 & 0.5807 & 0.6933 & 0.9899 & 0.4848 \\
    Sol-Attn & 0.9659 & 0.9571 & 0.5897 & 0.7030 & 0.9896 & 0.4303 \\
    \rowcolor{black!9}
    \textbf{QASK-Edit} & 0.9665 & 0.9551 & 0.5891 & 0.7072 & 0.9901 & 0.4515 \\
    \bottomrule
  \end{tabularx}
  \endgroup
\end{table*}

\begin{table*}[t]
  \centering
  \caption{Raw VBench metrics on VACE-R2V, rounded to four decimals without percentage scaling. These per-dimension values precede the normalization used to compute the aggregate VACE-R2V score.}
  \label{tab:suppl_vace_vbench}
  \begingroup
  \footnotesize
  \setlength{\tabcolsep}{3pt}
  \renewcommand{\arraystretch}{1.16}
  \begin{tabularx}{\textwidth}{@{}l*{8}{>{\raggedleft\arraybackslash}X}@{}}
    \toprule
    Method & \shortstack{Subject\\consist.} & \shortstack{Background\\consist.} & \shortstack{Aesthetic\\quality} & \shortstack{Imaging\\quality} & \shortstack{Motion\\smooth.} & \shortstack{Dynamic\\degree} & \shortstack{Temporal\\flickering} & \shortstack{Overall\\consist.} \\
    \midrule
    \multicolumn{9}{@{}l}{\textit{Wan2.1-VACE-1.3B}} \\[2pt]
    Dense & 0.9869 & 0.9809 & 0.6018 & 0.6915 & 0.9957 & 0.0750 & 0.9957 & 0.2670 \\
    SVG2 & 0.9807 & 0.9745 & 0.6077 & 0.6915 & 0.9956 & 0.0500 & 0.9953 & 0.2661 \\
    SpargeAttn & 0.9822 & 0.9774 & 0.6090 & 0.6771 & 0.9954 & 0.1000 & 0.9946 & 0.2749 \\
    Sol-Attn & 0.9862 & 0.9790 & 0.6072 & 0.6915 & 0.9956 & 0.0750 & 0.9955 & 0.2674 \\
    \rowcolor{black!9}
    \textbf{QASK-Edit} & 0.9853 & 0.9799 & 0.6080 & 0.6950 & 0.9956 & 0.0750 & 0.9955 & 0.2679 \\
    \midrule
    \multicolumn{9}{@{}l}{\textit{Phantom}} \\[2pt]
    Dense & 0.9881 & 0.9768 & 0.6104 & 0.7047 & 0.9928 & 0.1500 & 0.9876 & 0.2515 \\
    SVG2 & 0.9836 & 0.9708 & 0.6078 & 0.7019 & 0.9923 & 0.1500 & 0.9868 & 0.2513 \\
    SpargeAttn & 0.9878 & 0.9736 & 0.6056 & 0.7072 & 0.9929 & 0.2500 & 0.9877 & 0.2575 \\
    Sol-Attn & 0.9872 & 0.9762 & 0.6129 & 0.7028 & 0.9926 & 0.1500 & 0.9872 & 0.2514 \\
    \rowcolor{black!9}
    \textbf{QASK-Edit} & 0.9878 & 0.9777 & 0.6110 & 0.7028 & 0.9929 & 0.1500 & 0.9873 & 0.2523 \\
    \bottomrule
  \end{tabularx}
  \endgroup
\end{table*}

\begin{table*}[t]
  \centering
  \caption{Auxiliary reference-assisted cosine similarities on OpenS2V-Eval and VACE-R2V. CLIP-I, DINO-I, and CLIP-T denote the CLIP image, DINO image, and CLIP text cosine-similarity fields, respectively. Values retain their original scale and are rounded to four decimals.}
  \label{tab:suppl_reference_similarity}
  \begingroup
  \footnotesize
  \setlength{\tabcolsep}{3pt}
  \renewcommand{\arraystretch}{1.16}
  \begin{tabularx}{\textwidth}{@{}l*{6}{>{\raggedleft\arraybackslash}X}@{}}
    \toprule
    & \multicolumn{3}{c}{OpenS2V-Eval} & \multicolumn{3}{c}{VACE-R2V} \\
    \cmidrule(lr){2-4}\cmidrule(l){5-7}
    Method & CLIP-I & DINO-I & CLIP-T & CLIP-I & DINO-I & CLIP-T \\
    \midrule
    \multicolumn{7}{@{}l}{\textit{Wan2.1-VACE-1.3B}} \\[2pt]
    Dense & 0.6130 & 0.3611 & 0.3507 & 0.8057 & 0.5616 & 0.3475 \\
    SVG2 & 0.6064 & 0.3524 & 0.3513 & 0.8028 & 0.5494 & 0.3491 \\
    SpargeAttn & 0.5700 & 0.3173 & 0.3570 & 0.7522 & 0.5268 & 0.3538 \\
    Sol-Attn & 0.6034 & 0.3485 & 0.3520 & 0.8009 & 0.5486 & 0.3496 \\
    \rowcolor{black!9}
    \textbf{QASK-Edit} & 0.6158 & 0.3550 & 0.3517 & 0.8033 & 0.5485 & 0.3509 \\
    \midrule
    \multicolumn{7}{@{}l}{\textit{Phantom}} \\[2pt]
    Dense & 0.6382 & 0.3991 & 0.3447 & 0.8121 & 0.6325 & 0.3356 \\
    SVG2 & 0.6319 & 0.3917 & 0.3461 & 0.8099 & 0.6223 & 0.3362 \\
    SpargeAttn & 0.5934 & 0.3491 & 0.3470 & 0.7623 & 0.5575 & 0.3397 \\
    Sol-Attn & 0.6383 & 0.3948 & 0.3464 & 0.8128 & 0.6301 & 0.3361 \\
    \rowcolor{black!9}
    \textbf{QASK-Edit} & 0.6467 & 0.4007 & 0.3452 & 0.8153 & 0.6362 & 0.3381 \\
    \bottomrule
  \end{tabularx}
  \endgroup
\end{table*}

\subsection{FiVE-Bench Annotation Revisions}
\label{app:five_annotation_revisions}

To clarify the intended scene after object removal and the object to be edited, we adopt the \texttt{better-remove} annotations provided by StreamEdit~\cite{jiao2026streamedit} for the \(10\) object-removal tasks in the sixth category of FiVE-Bench. The revisions cover three aspects. First, target prompts replace negative removal instructions with affirmative descriptions of the intended scene and remove residual action descriptions associated with the object to be removed. Second, source-object annotations identify the specific object to be removed. Third, target-object annotations specify the background or content that should appear after removal.

For example, when removing a cat's toy, the source-object annotation changes from ``cat'' to ``toy,'' and the target-object annotation changes from ``toy'' to ``cozy living room.'' The revised target prompt directly describes a cat pouncing in the living room. In total, these revisions modify \(10\) target prompts, \(8\) source-object annotations, and \(10\) target-object annotations. The source videos, source prompts, editing instructions, and evaluation questions remain unchanged for all samples. The \(409\) editing tasks in the other five categories are also unchanged. We use the revised annotations consistently across all compared methods.

\section{Complete Generalization Results}
\label{app:generalization_full}

We provide the full metric breakdown for Wan2.1-VACE-1.3B~\cite{jiang2025vace} and Phantom~\cite{liu2025phantom} on OpenS2V-Eval and the reference-to-video (R2V) subset of VACE-Benchmark. Each model--method pair is evaluated on \(330\) OpenS2V videos and \(40\) VACE-R2V videos. All generalization experiments use a resolution of \(832\times480\) and a fixed length of \(81\) frames.

\paragraph{Aggregate scores}
Table~\ref{tab:suppl_gen_scores} reports the Open, Single, and Human Total Scores from OpenS2V-Eval and the VACE-R2V Normalized Average. Open is the open-domain score reported in the main paper. Single and Human denote the evaluator's additional tracks. All aggregate quality scores are reported on a \(0\)--\(100\) scale. The VACE-R2V score follows the benchmark's normalization and aggregation of VBench~\cite{huang2024vbench} metrics.

\paragraph{Metric breakdown}
Tables~\ref{tab:suppl_opens2v_vbench} and~\ref{tab:suppl_vace_vbench} report the raw VBench outputs before normalization or conversion to a percentage scale. Table~\ref{tab:suppl_reference_similarity} additionally reports the reference-assisted CLIP image, DINO image, and CLIP text cosine similarities recorded by the evaluation pipeline. These auxiliary similarities are listed separately from the benchmark aggregate scores. Dynamic degree measures the amount of motion.

\paragraph{Results}
Relative to Dense, {QASK-Edit} improves the OpenS2V open-domain score by \(1.01\) points on Wan2.1-VACE-1.3B and \(1.22\) points on Phantom. Its VACE-R2V normalized score remains close to Dense on both backbones. The overall end-to-end speedups are \(1.36\times\) for Wan2.1-VACE-1.3B and \(1.32\times\) for Phantom.

\end{document}